%% file: main.tex
\documentclass[runningheads]{llncs}
\usepackage{graphicx}

\usepackage{verbatim}

\usepackage{float}
\usepackage{subcaption}
\usepackage{adjustbox}
\usepackage[labelfont=bf,textfont=bf]{caption}
\usepackage[ruled,vlined,linesnumbered]{algorithm2e}
\usepackage{cite}
\usepackage{amsmath,amssymb,amsfonts}
\usepackage{textcomp}
\usepackage{xcolor}
\usepackage{todonotes}

\begin{document}
\title{Multiple Fairness and Cardinality constraints for Students-Topics Grouping Problem}
\titlerunning{Multiple Fairness and Cardinality constraints}
%
\author{Tai Le Quy\inst{1}\orcidID{0000-0001-8512-5854} \and
Gunnar Friege\inst{2}\orcidID{0000-0003-3878-9230}  \and
Eirini Ntoutsi\inst{3}\orcidID{0000-0001-5729-1003} }
\authorrunning{T. Le Quy et al.}
%
\institute{L3S Research Center, Leibniz University Hannover, Hanover, Germany \\
\email{tai@l3s.de}\\
\and
Institute for Didactics of Mathematics and Physics, Leibniz University Hannover, Hanover, Germany\\
\email{friege@idmp.uni-hannover.de}
\and 
Institute of Computer Science, Free University Berlin, Berlin, Germany\\
\email{eirini.ntoutsi@fu-berlin.de}}

\maketitle              
\begin{abstract}
Group work is a prevalent activity in educational settings, where students are often divided into topic-specific groups based on their preferences. The grouping should reflect the students' aspirations as much as possible. Usually, the resulting groups should also be balanced in terms of protected attributes like gender or race since studies indicate that students might learn better in a diverse group. Moreover, balancing the group cardinalities is also an essential requirement for fair workload distribution across the groups. In this paper, we introduce the \emph{multi-fair capacitated} (MFC) grouping problem that fairly partitions students into non-overlapping groups while ensuring balanced group cardinalities (with a lower bound and an upper bound), and maximizing the diversity of members in terms of protected attributes. We propose two approaches: a heuristic method and a knapsack-based method to obtain the MFC grouping. The experiments on a real dataset and a semi-synthetic dataset show that our proposed methods can satisfy students' preferences well and deliver balanced and diverse groups regarding cardinality and the protected attribute, respectively. 
 
\vspace{-10pt}
\end{abstract}

\input{introduction}
\input{relatedwork}
\input{problem}
\input{methodology}

\input{evaluation}
\vspace{-5pt}
\section{Conclusion and outlooks}
\label{sec:conclusion}
\vspace{-5pt}
In this work, we introduced the MFC grouping problem that ensures fairness in multiple aspects. We aim to ensure fairness 1) in terms of assignment by maximizing student satisfaction and 2) a fair-representation of students in each group according to the protected attributes like gender or race. In parallel, we balance the cardinality of the resulting groups with a lower and an upper bound. We implemented two proposed methods for the MFC grouping problem: the heuristic approach that prioritizes the students' preferences in the assignment, whereas the knapsack-based approach takes into account the students' preferences and the cardinality of the groups during the assignment step, which is formulated as a maximal knapsack problem. Our experiments show that our methods are effective regarding student satisfaction and fairness w.r.t the protected attribute while maintaining the balance in cardinality with the given bounds. In the future, we plan to extend our approach to more than one protected attribute, such as gender and race, as well as to further investigate the groups' characteristics w.r.t. students' abilities, communicative skills, etc., and other definitions with different aspects of fairness in the educational environment. 
\vspace{-10pt}
\section*{Acknowledgment}
\vspace{-5pt}
The work of the first author is supported by the Ministry of Science and Culture of Lower Saxony, Germany, within the Ph.D. program ``LernMINT: Data-assisted teaching in the MINT subjects”. 
%
%
\vspace{-10pt}
\bibliographystyle{splncs04}
\bibliography{bibliography}
\vspace{-5pt}
\end{document}

%% file: introduction.tex
\section{Introduction}
\label{sec:introduction}

Teamwork plays a vital role in educational activities, as students can work together to achieve shared learning goals. By working in groups, students have better communication and become more social and creative. Moreover, they can learn about leadership, higher-order thinking, conflict management \cite{hansen2006benefits,ford2003fair} etc.
A common practice for group work is as follows: The instructor provides a list of topics, projects, or tasks\footnote{We use the term ``topic'' to refer to all of them.} according to which the different non-overlapping groups of students should be formed. The grouping procedure can be performed randomly or based on students' preferences~\cite{miles1998fairness} typically expressed as a ranking over the provided topics. Or, the teacher just says: ``Find yourself into groups"; this case is not random and not according to preferences with regard to the topic but is triggered by my social networks or is due to the fact of how sitting in the neighborhood. The important case and often in educational settings, is the grouping with regard to preferences. In the classroom, this costs time, and a suitable algorithm could help. Therefore, in this work, we consider the case of grouping w.r.t. students' preferences. 

The grouping process should consider a variety of requirements.
First, students' preferences should be taken into account (i.e., \emph{student satisfaction requirement}). A grouping is considered satisfactory if it can satisfy the students' preferences as much as possible. It is important to give students equal opportunity to obtain their preferred topics \cite{magnanti2018allocating}.
Second, the groups should be balanced in terms of their cardinalities, so all students share a similar amount of work (i.e., \emph{group cardinality requirement}) because when groups have unequal sizes, and the minority group is smaller than a critical size, the minority cohesion widens inequality \cite{oliveira2022group}.
Third, the instructor might be interested in fair-representation groups w.r.t. some protected attributes like gender or race~\cite{krass2006university} (i.e., \emph{group fairness requirement}), because a higher female ratio in the groups may have a positive effect on the groups' performance \cite{fenwick2001effect}.

These requirements have been already discussed in the related work but are typically treated independently. For example, 
fairness has been discussed in the context of group assignments \cite{ford2003fair}, assignment of group members to tasks \cite{miles1998fairness} or students to projects \cite{rezaeinia2021efficiency}. Student satisfaction requirement is formulated by the sum of the number of topics staffed \cite{lopes2008optimization} or the sum of the utilities of the topics assigned to students based on the ranking of preferences chosen by students \cite{magnanti2018allocating}. The resulting group cardinality can be satisfied by the heuristic method \cite{mulvey1984solving}, or the hierarchical clustering approach \cite{le2021fair}.
However, providing a grouping solution that simultaneously satisfies all these constraints is hard. And, ``in general, it is not possible to assign all students to their most preferred project'' \cite{rezaeinia2021efficiency}.

In this work, we introduce the problem of \emph{multi-fair capacitated (MFC) grouping} that aims to ensure fairness of the resulting groups in multiple aspects. We target fairness in terms of assignment (maximizing the student satisfaction by the objective function) and fairness w.r.t. protected attributes as well balanced cardinalities across the resulting groups (with a lower bound and an upper bound). We define two fairness constraints of the grouping based on the Nash social welfare notation \cite{nash1950bargaining}, and the balance score \cite{chierichetti2017fair}. We propose two approaches to solve the MFC grouping problem. The first is a heuristic, whereas, in the second, we reformulate the assignment step as a maximal knapsack problem.

The rest of our paper is structured as follows: we overview the related work in Section \ref{sec:relatedwork}. The multi-fair capacitated grouping problem is introduced in Section \ref{sec:problem}.
Section \ref{sec:methodology} presents the solutions to the MFC problem. The experimental evaluation on several educational datasets is described in Section~\ref{sec:evaluation}. Finally, section \ref{sec:conclusion} summarizes our conclusions and outlook.

%% file: relatedwork.tex
\section{Related work}
\label{sec:relatedwork}
\vspace{-5pt}
In the education domain, Miles et al. \cite{miles1998fairness} investigated the problem of assignment of group members to tasks. They examined the viability of four methods to assign students into the groups: random, ability, personal influence, and personal influence with justification. Concerning a diversity of features such as skills, genders, and academic backgrounds, Krass et al. \cite{krass2006university} investigated the problem of assigning students to multiple non-overlapping groups. The problem was solved by an integer programming model to minimize the number of overlaps. In a similar research \cite{cutshall2007indiana}, the authors assign students into groups based on their academic background and gender. However, students' preferences were not considered.

To consider both efficiency and fairness, Magnanti et al. \cite{magnanti2018allocating} solve an integer programming formulation with two objectives: maximizing the total utility computed by the rank of student's preferences (efficiency) and minimizing the number of students assigned to the projects which they do not prefer (fairness). Recently,  Rezaeinia et al. \cite{rezaeinia2021efficiency} introduced a lexicographic approach to prioritize the goals. The efficiency objective is computed based on the utility, similar to \cite{magnanti2018allocating}; however, the authors adapted the Jain's index \cite{jain1984quantitative} to measure the fairness of the assignment.

Related to our grouping problem is the problem of assigning reviewers to papers~\cite{hartvigsen1999conference,long2013good,
stelmakh2021peerreview4all}. However, in the paper-reviewer assignment problem, each reviewer should be assigned several papers, and each paper should be assigned several reviewers \cite{hartvigsen1999conference}. Meanwhile, in the students grouping problem, we attempt to generate non-overlapping groups of students \cite{krass2006university}, and each student can be assigned to only one group \cite{rezaeinia2021efficiency}.

The knapsack problem formulation has been used for finding good clustering assignments~\cite{le2021fair}. However, the minimum capacity of a group (cluster) is not ensured. Recently, Stahl et al.~\cite{stahl2016fair} introduced a fair knapsack model to balance the price given by the data provider and the suggested price of the customer. The data vendors propose the data for an \textit{ask price}, and customers can negotiate a \textit{bid price}. The data quality is adjusted to satisfy the price bargained by the customer and ensure the final selling price is fair. Next, Fluschnik et al.~\cite{fluschnik2019fair} proposed three concepts of fair knapsack (individually best, diverse and fair knapsack) to solve the problem of choosing a subset of items with the total cost is not greater than a given \textit{budget} while taking into account the preferences of the voters. 

The Nash social welfare (Nash equilibrium) \cite{nash1950bargaining} was used as the solution concept for fairness \cite{fluschnik2019fair}, i.e., fairness is ensured by the objective function. The group fairness definition for the knapsack problem was investigated recently by Patel et al.~\cite{patel2021group}. Fairness is defined by several constraints. In their study, each item is characterized by a \textit{category},  their goal is to select a subset of items such that the total value of the selected items is maximized, and the total weight does not surpass a given weight while each category is \textit{fairly} represented. The notion of \textit{group fairness} is defined based on three criteria (the number of items, the total value of items, and the total weight of items in each category).

In this work, we introduce the MFC grouping problem that ensures fairness in multiple aspects. In particular, we guarantee fairness in terms of topics assignment (by objective function) in parallel with cardinality (lower bound and upper bound on the group cardinality) and fairness w.r.t protected attributes (by prime of constraints). To the best of our knowledge, the proposed problem has not been studied before and, as already discussed, comprises a useful tool to ensure fairness in educational activities.
\vspace{-5pt}

%% file: problem.tex
\section{Problem definition}
\label{sec:problem}
Let $X = \{x_1, x_2,\cdots, x_n\}$ is a set of $n$ students, and $T = \{t_1, t_2,\cdots, t_m\}$ is a set of $m$ topics. For an integer $n$ we use [n] to denote the set $\{1, 2, \cdots,n\}$. Each student can choose $h$ topics as their preferences. We store the preferences of student in a matrix, namely $wishes$. In which, each row $wishes_{i}$ contains the list of $h$ topics that are preferred by student $i$. We use the matrix $V$ to record the student's level of interest in the topics. The preference of topic $t_j$ chosen by student $x_i$ is represented by a number $v_{ij}$; the most interested topic of student $x_i$ is expressed by the highest value of $v_{ij}$. Likewise, each topic $t_j$ can be chosen by several students. A priority matrix $W$ containing the value computed based on the time when the students register. The value $w_{ij}$ represents the registration of student $x_i$ on the topic $t_j$. In which, the first register of topic $j$ leads to the highest value of $w_{ij}$. If the topic $t_j$ is not chosen by student $x_i$ then $v_{ij} = 0$ and $w_{ij} = 0$. 

\begin{figure*}[!h]
    \centering
    \vspace{-15pt}
    \includegraphics[width=\textwidth]{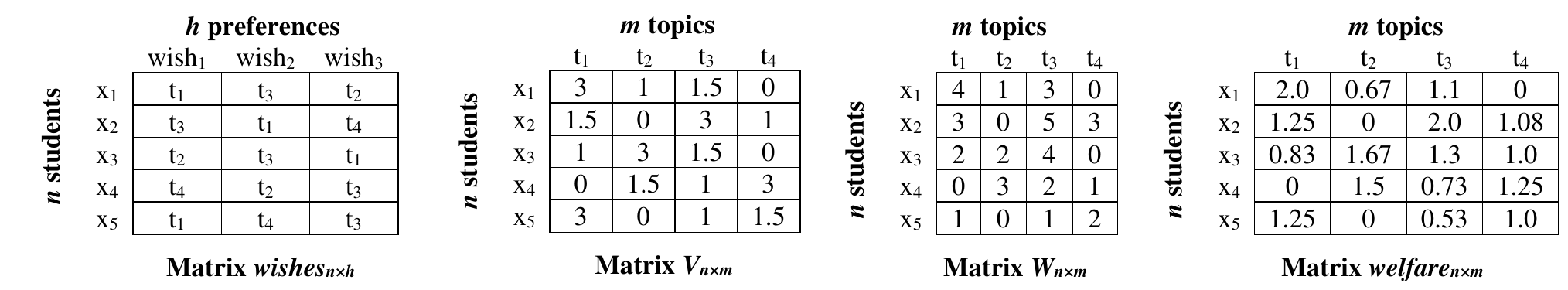}
    \caption{An example of matrices $wishes$, $V$, $W$ and $welfare$ of a dataset with 5 students and 4 topics}
    \vspace{-10pt}
    \label{fig:matrix_example}
\end{figure*}

Let $\psi: V \times W \rightarrow \mathbb{R}$ be the aggregate function of matrices $V$ and $W$. For each student $x_i$, we define a \emph{welfare} value: $welfare_{ij}=\psi(v_{ij},w_{ij})$. 
\begin{equation}
\label{eq:aggregate_function}
    \psi(v_{ij},w_{ij}) = \alpha v_{ij} + \beta w_{ij}
\end{equation}
In Eq. \ref{eq:aggregate_function}, $\alpha$ and $\beta$ are the parameters indicating the weight of each component. Figure \ref{fig:matrix_example} illustrates an example of matrices $wishes$, $V$, $W$ and $welfare$ of a dataset with 5 students and 4 topics. Matrix $V$ is computed based on the \emph{preferences} of students. In details, $V_{i,wishes_{ip}} = h/p$ with $p \in [h]$, where $p$ indicates the order of preferences. The matrix $welfare$ is computed by Eq. \ref{eq:aggregate_function} with $\alpha = 1$ and $\beta = 1$.

\vspace{-5pt}
The goal of a grouping problem is to distribute $n$ students into $k$ disjoint groups $\mathcal{G} = \{G_1, G_2,\cdots, G_k\}$, where $k \leq m$, that maximizes the students' preferences w.r.t the registration time, formulated by the objective function:
\begin{equation}
\label{eq:objective_function}
    \mathcal{L}(X,\mathcal{G}) = \prod_{r=1}^{k}\sum_{i=1}^{n}welfare_{ij_r} * y_{ij_r}
\end{equation}
In other words, the goal is maximize product of the total \emph{welfare} obtained from each group $G_i$. In Eq. \ref{eq:objective_function}, a set of indexes $J = \{j_1, j_2, \cdots, j_k\}$ of $k$ selected topics is defined as $J = \{j | x_i \in G_r, welfare_{ij}>0\}$, $\forall r \in [k]$. Variable $y_{ij_r}$ is the flag of $x_i$, where $y_{ij_r} = 1$ if $x_i$ is assigned to the group of topic $t_{j_r}$, otherwise $y_{ij_r}= 0$. 

Similar to \cite{fluschnik2019fair}\footnote{In \cite{fluschnik2019fair}, the Nash equilibrium was defined as $\prod_{v_i 
\in V}(1+\sum_{a\in S}u_{i}(a))$ (the typical formula is $\prod_{v_i 
\in V}\sum_{a\in S}u_{i}(a)$, where $v_i$ is a voter in a set of voters $V$, $a$ is an item of the knapsack $S$, and $u_{i}(a)$ represents the extent to which $v_i$ enjoys $a$. The knapsack $S$ is fair if that product is maximized.}, Eq.~\ref{eq:objective_function} is the representation of the Nash social welfare (Nash equilibrium)~\cite{nash1950bargaining} function
therefore, we can call a grouping is satisfactory if it maximizes the product in the objective function $\mathcal{L}(X,\mathcal{G})$. Furthermore, we add one to the sum $\sum_{i=1}^{n}welfare_{ij_r} * y_{ij_r}$ to avoid the phenomenon that the sum of welfare in a certain group might be zero. The objective function $\mathcal{L}(X,\mathcal{G})$ is rewritten as follows:

\begin{equation}
\label{eq:objective_function_new}
    \mathcal{L}(X,\mathcal{G}) = \prod_{r=1}^{k}(1+ \sum_{i=1}^{n}welfare_{ij_r} * y_{ij_r})
\end{equation}

\textbf{Fairness of grouping in terms of protected attributes:}
Assume that each student is characterized by a binary protected attribute $P=\{0, 1\}$, e.g., $gender = \{male, female\}$. Let $\varphi: X \rightarrow P$ denotes the demographic category to which the student belongs, i.e., male or female. Fairness of a group in terms of the balance score w.r.t. protected attribute~\cite{chierichetti2017fair} is defined as the minimum ratio between two categories.
\begin{equation}
\label{eq:balance_score}
balance(G_r)_{\forall G_r\in \mathcal{G}}=\min\left(\frac{|\{x \in G_r \mid \varphi(x)=0\}|}{|\{x \in G_r \mid \varphi(x)=1\}|},
\frac{|\{x \in G_r \mid \varphi(x)=1\}|}{|\{x \in G_r \mid \varphi(x)=0\}|}\right)
\end{equation}

Fairness of a grouping w.r.t the protected attribute is computed as:
\begin{equation}
\label{eq:balance_group}
    balance(\mathcal{G}) = \min_{\forall G_r \in \mathcal{G}} balance(G_r)
\end{equation}

\textbf{Capacitated grouping:} Being inspired of the capacitated clustering problem~\cite{mulvey1984solving}, we call a grouping is \textit{capacitated} if the cardinality of each group $G_r$, i.e., $|G_r|$, is between a given lower bound $C^l \geq 0$ and an upper bound $C^u$.

We now introduce the \textit{multi-fair capacitated} (MFC) grouping problem, which satisfies the capacity constraint and two fairness constraints.
\begin{definition}
\label{def:assignment}
{\rm \textbf{MFC grouping problem}

We describe the MFC problem as finding a grouping $\mathcal{G} = \{G_1, G_2,\cdots, G_k\}$ that distributes a set of students $X$ into $k$ groups corresponding to $k$ topics, and satisfies the following constraints: \textit{1)} The assignment is fair, i.e., maximizing the objective function $\mathcal{L}(X,\mathcal{G})$ (see Eq. \ref{eq:objective_function_new}); \textit{2)} The balance of each group $G_r$ is maximized, i.e., the fairness constraint w.r.t the protected attribute (see Eq. \ref{eq:balance_group}); \textit{3)} The cardinality of each group $G_r \in \mathcal{G}$ is in between the lower bound $C^l$ and the upper bound $C^u$ (the cardinality constraint).
}
\end{definition}

%% file: methodology.tex
\section{Methodology}
\label{sec:methodology}
In this section, we propose two approaches to solve the MFC grouping problem. The former is based on a heuristic approach (Section \ref{subsec:heuristic}) while the latter is the reformulation of a knapsack problem (Section \ref{subsec:knapsack}).
\subsection{A heuristic approach}
\label{subsec:heuristic}
The main idea of our heuristic method is to assign a student to the topic which is the highest favorite one of the student's preferences. This approach is divided into two main phases, as presented in the Algorithm \ref{alg:heuristic}.

In the first step, we maximize the students' preferences by assigning them to the group with the highest priority on their desires. We consider each student and each preference accordingly (lines 30, 31). If many students choose the same topic, we will assign the current observed student to the topic if that student has the highest $welfare$ value. Moreover, the capacity of groups is also taken into account during the assignment procedure (lines 32, 33). 

In the second step, we will adjust the assignment to satisfy the constraints (function \emph{GroupAdjustment}). If there are any ungrouped students, we will try to assign them to the existing groups (line 2 to line 8). If all groups are full, we will choose the most prevalent topic preferred by the remaining ungrouped students and then assign them to such a topic (line 9 to line 16). The cardinality constraint is satisfied in the next step with some modifications to the groups' members. We disband groups that have too few students and assign such ungrouped ones to other groups. This procedure is repeated until all groups have the desired capacity (line 18 to line 25).

\textbf{Computational complexity}: The computational time of step 1 of the Algorithm \ref{alg:heuristic} is $\mathcal{O}(n \times h)$ while the processing stage for students who have not been assigned groups costs $\mathcal{O}(n \times m)$. In step 2, the group adjustment phase, the maximum running time is $\mathcal{O}(C^l\times n\times m)$ because the algorithm has to deal with every group having cardinality less than $C^l$. In short, because $C^l \ll n$, we can conclude the computational complexity of the algorithm is $\mathcal{O}(n\times m)$, where $n$ is the number of students and $m$ is the number of topics.

\subsection{Knapsack-based approach}
\label{subsec:knapsack}
In the heuristic approach, we tend to assign students to their most favorite topics. This assignment can be detrimental in satisfying the preferences of other students because some of the remaining students will no longer have a topic to be assigned to even though they also have a high degree of interest in that topic. Therefore, assigning students to their second or third favorite topic could improve student satisfaction overall. Hence, we propose a new approach whereby we will search for the most suitable students for each topic. We will formulate the task of selecting the ``best'' students for a group of the MFC grouping problem as a \emph{maximal knapsack} problem \cite{mathews1896partition}.

\begin{algorithm}[!htb]
\SetAlgoLined
\KwIn{$\mathcal{X} = \{x_1, x_2, \ldots, x_n\}$: a set of students \newline
$n$: the number of students;
$h$: the number of preferences;
$m$: the number of topics \newline
$C^l, C^u$: the given minimum and maximum capacity of each group\newline
$wishes_{n \times h}$: the preferences of students \newline
$V_{n \times m}$: a students' level of interest matrix \newline
$W_{n \times m}$: a priority matrix
}
\KwOut{A grouping with $k$ groups}

\SetKwFunction{FAdjustment}{\textbf{GroupAdjustment}}
\SetKwProg{Fn}{Function}{:}{}
  \Fn{\FAdjustment{$groups$}}{
        \For{$i\gets1$ \KwTo $n$ } 
    {   
        \For{$l\gets1$ \KwTo $m$ }
        {  
            \If {($i\notin groups[l]$) and ($len(groups[l] < C^l$))}
            {
                $groups$[$l$].append($i$)\;
            }
        }
    }
    \While{$len(unassigned\_students) > 0$}{
    $id \gets$ the most prevalent topic desired by remaining students\; 
    \For{$i \in unassigned\_students$}
    {
        \If{$len(groups[id] < C^u$))}{$groups$[$id$].append($i$)\; }
    }
    }

    $n\_items \gets 1$\;  
    \While{(cardinalities of all groups $\notin$ $[C^l, C^u]$)}
    {
    \If{$n\_items < C^l$}{
    Resolve the groups with cardinality $n\_items$\; 
    Assign ungrouped students to the remaining groups having cardinality  $< C^u$\;
    $n\_items++$\;    
    }
    }
    
    \KwRet $groups$\;
  }
\SetKwFunction{FMain}{\textbf{main}} 
\label{alg:main_function}
\SetKwProg{Fm}{Function}{:}{}
  \Fm{\FMain{}}{
//Step 1: Assign students to groups

$groups \gets \emptyset$\;
$welfare \gets \psi(V,W)$ //Eq. \ref{eq:aggregate_function}\;
\For{$i\gets1$ \KwTo $n$}
    {
        \For{$j\gets1$ \KwTo $h$}
        {
            \If {(topic $wishes[i,j]$ is the most preferred topic of student $i$) and ($welfare_{i,wish[i,j]}$ is the highest value among students choosing topic $wishes[i,j]$) and ($len(groups[wishes[i,j]] < C^l$))}
            {
                $groups$[$wishes$[$i,j$]].append($i$)\;
            }
        }
    }
    
//Step 2: Adjustment

GroupAdjustment($groups$)\;

}
\Return{$groups$}\;
\caption{Heuristic algorithm for the MFC grouping problem}
\label{alg:heuristic}
\end{algorithm}
\clearpage

Let $capacity$ is a cardinality array, $capacity_i = |x_i|$; $welfare_{ij} = \psi(v_{ij},w_{ij})$ and the indexes of $k$ topics $J=\{j_1, j_2,\cdots, j_k\}$ will be chosen for the resulting groups. For each topic $t_{j_r} \in T$, $\forall r \in [k]$, i.e., $r$ is the index of the selected knapsack, the goal is to select a subset of students ($G_r$), such that:
\begin{equation}
\label{eq:knapsack}
\begin{aligned}
    \textrm{maximize}  \sum_{i=1}^{n} welfare_{ij_{r}}*y_{ij_{r}}
    \phantom{13233356431}\\
    \textrm{subject to} \quad 
    \begin{cases}
    \sum_{i=1}^{n} capacity_{i}*y_{ij_{r}} \leq C^u ~\text{or}\\ \sum_{i=1}^{n} capacity_{i}*y_{ij_{r}} \leq C^l\\
    balance(G_r) \text{ is maximized}
\end{cases}
\end{aligned}
\end{equation}

where
\[
    y_{ij_r}= 
\begin{cases}
    1,& \text{if student } x_i \text{ is assigned to topic } t_{j_r}\\
    0,              & \text{otherwise}
\end{cases}
, \forall r \in [k]
\]

We formulate Eq. \ref{eq:knapsack} as a maximal knapsack problem. In the knapsack problem, the goal is to find a set of items that maximizes the total value, and the total weight is less than or equal to a given limit. In our case, for each selected topic, we find a set of students that maximizes the total $welfare$, while the total $capacity$, i.e., the group cardinality, is in the range of the given bounds.

The pseudo-code of our knapsack-based method is described in Algorithm \ref{alg:knapsack}, which is a two-phase approach. In the first step, for each topic, we find the most suitable candidates among the unassigned students by the result of a vanilla knapsack problem \cite{mathews1896partition}. To solve the maximal knapsack problem (Eq. \ref{eq:knapsack}), we use the dynamic programming to get the result which is a group of students having the maximum $welfare$ and the cardinality is in the range $[C^l, C^u]$ (line 13 to line 18). The group adjustment step is similar to Algorithm \ref{alg:heuristic}, which performs a fine-tuning phase in the grouping.

\textbf{Computational complexity}: 
In the first assignment step, most of the computational time is devoted to the knapsack problem, which costs $\mathcal{O}(n\times C^u)$ for each topic. Hence, the first assignment step consumes $\mathcal{O}(m \times n \times C^u)$. Similar to Algorithm \ref{alg:heuristic}, the maximum running time the group adjustment step is $\mathcal{O}(C^l\times n\times m)$. Because $C^l \ll n$ and $C^u \ll n$, the computational complexity of the algorithm is $\mathcal{O}(n\times m)$.

\begin{algorithm}[!ht]
\SetAlgoLined
\KwIn{$\mathcal{X} = \{x_1, x_2, \ldots, x_n\}$: a set of students \newline
$n$: the number of students;
$h$: the number of preferences;
$m$: the number of topics \newline
$C^l, C^u$: the given minimum and maximum capacity of each group\newline
$wishes_{n \times h}$: the preferences of students \newline
$V_{n \times m}$: a student's level of interest matrix\newline
$W_{n \times m}$: a priority matrix
}
\KwOut{A grouping with $k$ groups}

\SetKwFunction{FAdjustment}{GroupAdjustment}
\SetKwProg{Fn}{Function}{:}{}
  \Fn{\FAdjustment{$groups$}}{
    //The function is described in Algorithm \ref{alg:heuristic}
    
    \KwRet $groups$\;
  }
\SetKwFunction{FMain}{\textbf{main}} 
\label{alg:main_function}
\SetKwProg{Fm}{Function}{:}{}
  \Fm{\FMain{}}{
//Step 1: Assign students to groups based on a maximal knapsack method

$groups \gets \emptyset$\;
$welfare \gets \psi(V,W)$ //Eq. \ref{eq:aggregate_function}\;
\For{$id\gets1$ \KwTo $m$}
{
    $capacity \gets$ get\_capacity($unassigned\_students$)\; 
    $values \gets$ get\_welfare($unassigned\_students, welfare$)\;
    $n\_items \gets $ len($unassigned\_students$)\;
    \If{$n\_items > 0$}
    {
        \If{$n \bmod C^l = 0$ }
        {
            $selected\_points \gets$ knapsack($values, capacity, n, C^l$)\;
        }
        \Else
        {
            $selected\_points \gets$ knapsack($values, capacity, n, C^u$)\;
        }
        $groups[id] \gets  selected\_points$\;
    }
}
//Step 2: Adjustment

GroupAdjustment($groups$)\;
}
\Return{$groups$}\;
\caption{Knapsack-based algorithm for the MFC grouping problem}
\label{alg:knapsack}
\end{algorithm}
\vspace{-5pt}

%% file: evaluation.tex
\section{Evaluation}
\label{sec:evaluation}
In this section, we present our experiments and the performance of our proposed approaches on two educational datasets.
\subsection{Datasets}
\label{subsec:dataset}
We evaluate our proposed methods on two variations of a real dataset used often in educational data science~\cite{lequy2022survey} 
 and a real data science dataset collected at our institute. An overview of datasets is summarized in Table \ref{tbl:dataset}.
\vspace{-15pt}
\begin{table*}[!htb]
\centering
\caption{An overview of the datasets}\label{tbl:dataset}
\begin{tabular}{ccccc}
\hline
Dataset                              & \multicolumn{1}{l}{\#instances} & \multicolumn{1}{l}{\#attributes} & Protected attribute & Balance \\ \hline
Real data science & 24  & 23 & Gender (F: 8, M: 16) & 0.5 \\
Student - Mathematics & 395  & 33 & Gender (F: 208, M: 187) & 0.899 \\
Student - Portuguese  & 649  & 33 & Gender (F: 383; M: 266) & 0.695 \\
\hline
\end{tabular}
\vspace{-10pt}
\end{table*}

\textbf{Real data science dataset}\footnote{https://github.com/tailequy/tailequy.github.io/tree/main/fair-grouping/data}. This dataset is collected in a seminar on data science at our institute. Students have to register 3 desired topics out of 16 topics. The advisor will assign students into groups based on their preferences and the registration time. The data contain demographic information of students (attributes: \textit{ID, Name, Gender}) with their preferences (attributes: \textit{wish1, wish2, wish3}, registration time (attribute: \textit{Time}) and priority matrix \textit{W} which is represented by 16 attributes \textit{T1, \ldots, T16}.

\textbf{UCI Student performance dataset}\footnote{https://archive.ics.uci.edu/ml/datasets/Student+Performance}. This dataset consists of demographic, social, school-related attributes and students' grades in secondary education of two Portuguese schools in 2005 - 2006 \cite{cortez2008using} with two subjects: Mathematics and Portuguese. Because there is no given information about topics and preferences of students in the original dataset, we create a \emph{semi-synthetic} dataset by generating the preferences and the topics and merging them into the original version. The number of preferences \emph{h} and the number of topics \emph{m} are the main parameters of the data generator. 
For each student, we randomly generate $h$ different favorite topics, which are stored in the $h$ columns. Then, for each topic, we list the students who selected that topic and randomly generate (different) priorities for them. This matrix $W$ is stored in $m$ columns.
Therefore, the \emph{semi-synthetic} version will contain $(h+m)$ new attributes. 

\subsection{Experimental setups}
\label{subsec:experimental_setup}
\subsubsection{Parameter selection}
\label{subsubsec:parameters}
Similar to the settings of the real data science dataset, we set the number of wishes $h = 3$ for the UCI student performance dataset. Naturally, a group should contain at least 2 students; therefore, the number of topics is chosen to satisfy each group of 2 members assigned to a topic. Hence, we set $m = 200$ and $m = 325$ as the number of topics for the UCI student dataset - Mathematics and Portuguese subjects, respectively.
In addition, we set the parameters $\alpha = 1$ and $\beta = 1$ (Eq. \ref{eq:aggregate_function}), i.e., each component has the same weight.

\textbf{Parameters related to groups' cardinality}. Since the real data science is a very small dataset, our methods are evaluated with the lower bound $C^l$ in the range of $(2, \ldots, 8)$. For the UCI student performance dataset, we set $C^l = (2, \ldots, 18)$, as \cite{urbina2017associating} suggests that the average number of students per group should not exceed 20. The upper bound $C^u$ is set as $C^u = C^l + 1$ for all datasets.
\subsubsection{Evaluation measures}
\label{subsubsec:measurement}
We report our experimental results w.r.t. fairness in terms of grouping assignment, protected attribute and cardinality with the following measures:

\textbf{Nash equilibrium}. The Nash equilibrium is computed by the Eq. \ref{eq:objective_function_new}. However, the number of groups ($k$) is determined during the group assignment process, i.e., $k$ is different for the same set ${C^l, C^u}$, for each method. Hence, we normalize the Nash equilibrium of the final group assignment by the following logarithmic function:
\begin{equation}
\label{eq:nash_log}
    Nash = log_{k}\mathcal{L}(X,\mathcal{G})
\end{equation}

\textbf{Balance}. The fairness in terms of the protected attribute (Eq. \ref{eq:balance_group}).

\textbf{Satisfaction level of students' wishes}. It is measured by the ratio of the number of students who are satisfied, i.e., they are assigned to the topic of their preferences, out of the total number of students.
\begin{equation}
\label{eq:satisfaction}
    Satisfaction = \frac{\mid \{i|wishes_{ip} = k, i\in groups_k, p \in [h]\}\mid}{n}
\end{equation}

\subsection{Experimental results}
\label{subsec:results}
\subsubsection{Real data science}

\label{subsubsec:real_result}
\vspace{-10pt}
\begin{figure} [!htb]
     \centering
     \begin{subfigure}[b]{0.75\textwidth}
         \centering
         \includegraphics[width=\textwidth]{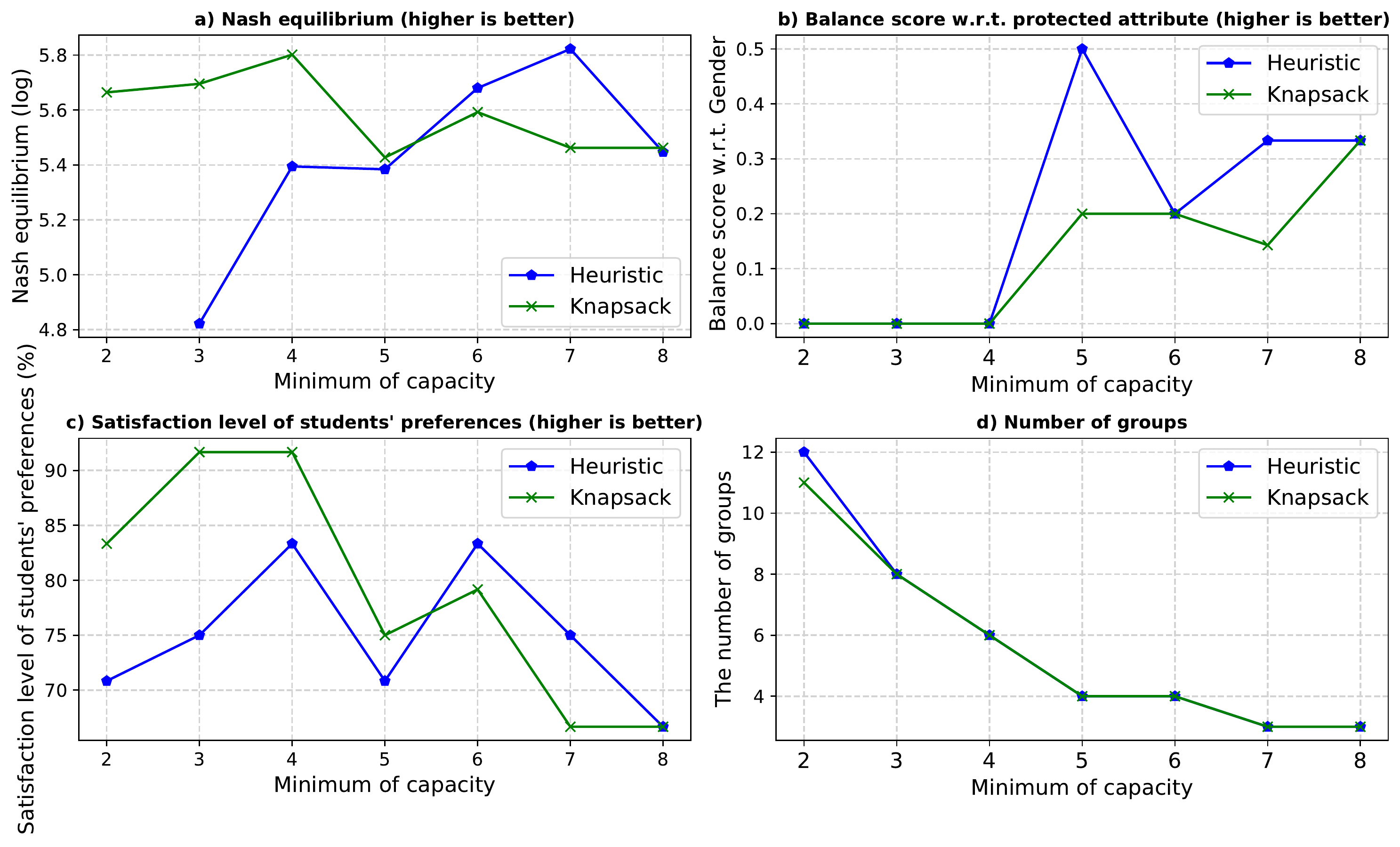}
         \label{fig:plot_real_data}
    \end{subfigure}
     \hfill
     \begin{subfigure}[b]{0.55\textwidth}
         \centering
         \vspace{-15pt}
         \includegraphics[width=\textwidth]{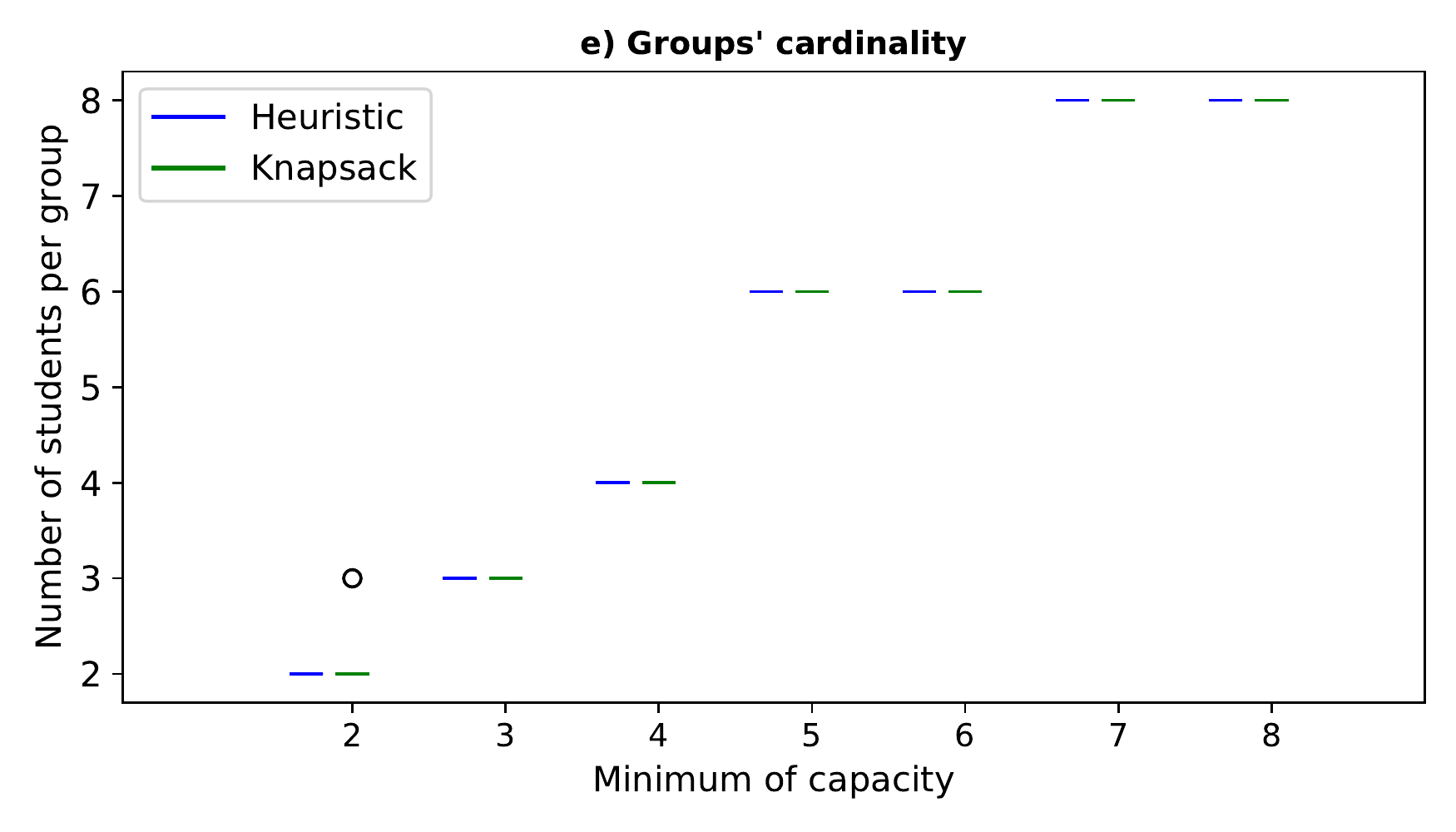}
         \label{fig:boxplot_real_data}
     \end{subfigure}
     \vspace{-15pt}
     \caption{Performance of different methods on the real data science dataset}
     \vspace{-15pt}
     \label{fig:real_data}
\end{figure}
\vspace{-5pt}
As demonstrated in Fig. \ref{fig:real_data}-a and Fig. \ref{fig:real_data}-b, the grouping results from the knapsack-based approach are better in terms of the Nash equilibrium. There are more students allocated to the groups as their preferences when the group's size is less than 6 persons. The satisfaction level decreases when the groups' cardinality increases. This is understandable since students have only a limited number of preferences (3 topics). When the group's cardinality increases, the desired topics become more diverse, and it is difficult to satisfy most students. In terms of fairness w.r.t. the protected attribute, the heuristic method outperforms the knapsack-based approach when there are at least 5 persons in a group (Fig. \ref{fig:real_data}-c). The number of groups and the group's cardinality are quite consistent for both methods, which are illustrated in Fig. \ref{fig:real_data}-d and Fig. \ref{fig:real_data}-e.

\subsubsection{UCI student performance - Mathematics}
\label{subsubsec:student_mat_result}
\begin{figure*} [!h]
     \centering
     \begin{subfigure}[b]{1\textwidth}
         \centering
         \includegraphics[width=\textwidth]{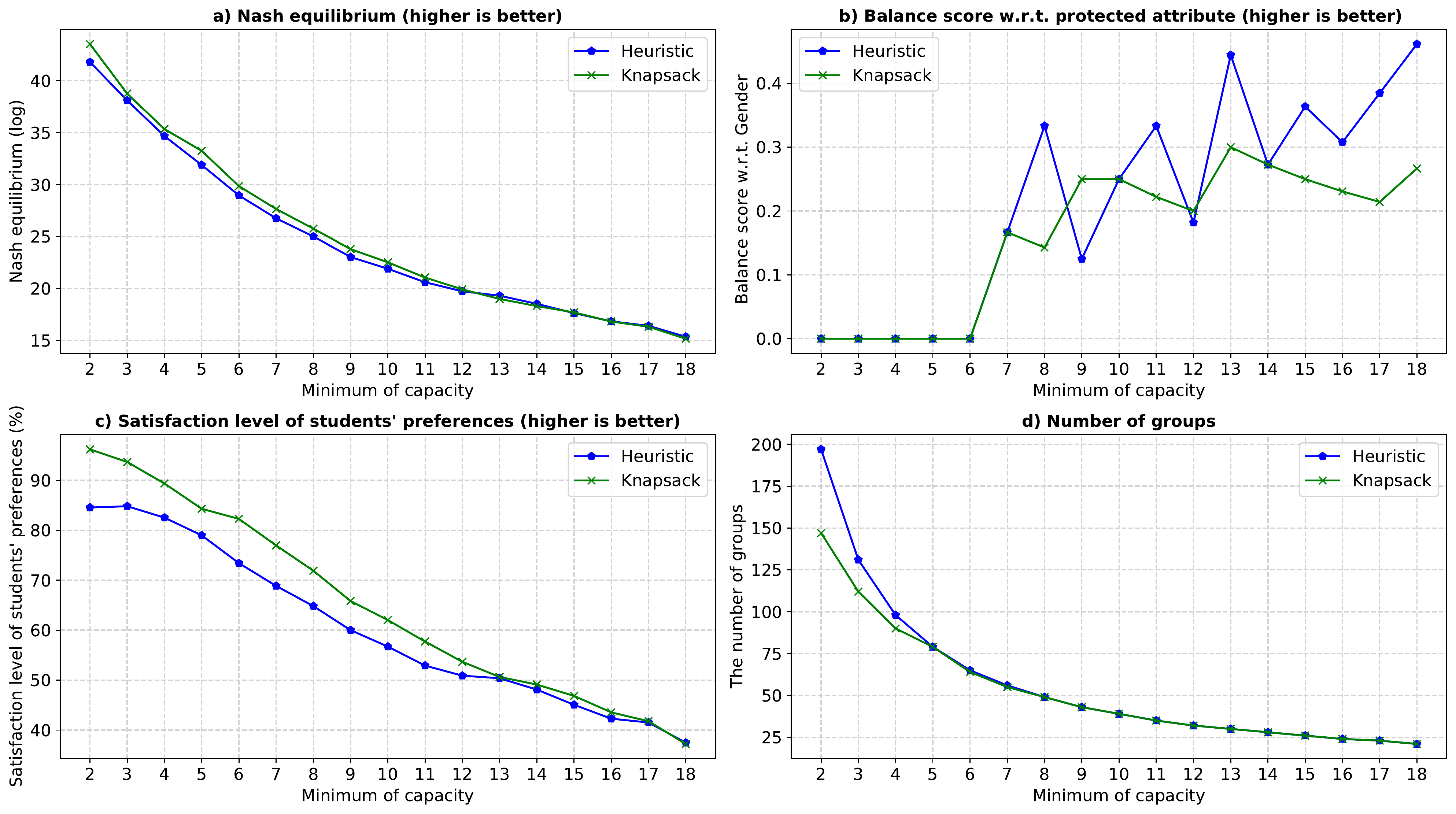}
         \label{fig:plot_student_mat}
    \end{subfigure}
     \hfill
     \begin{subfigure}[b]{0.9\textwidth}
         \centering
         \vspace{-13pt}
         \includegraphics[width=\textwidth]{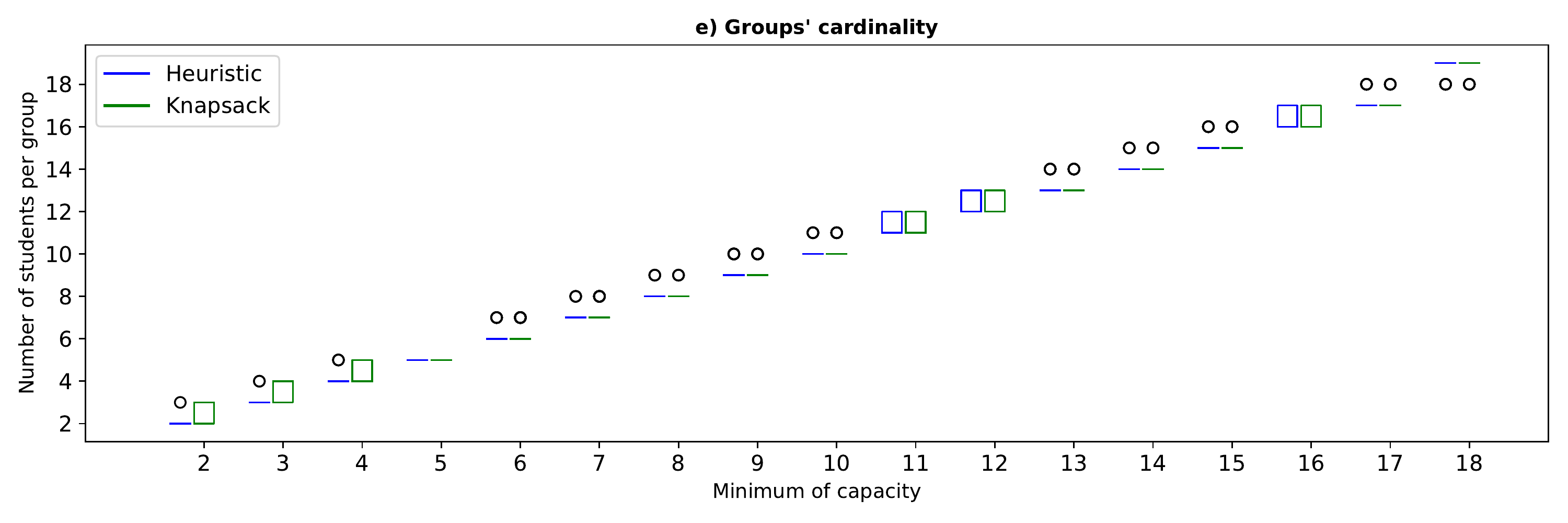}
         \label{fig:boxplot_student_mat}
     \end{subfigure}
     \vspace{-18pt}
     \caption{Performance of different methods on the UCI student performance dataset - Mathematics subject}
     \vspace{-15pt}
     \label{fig:student_mat}
\end{figure*}

The knapsack-based approach outperforms the heuristic method in terms of both Nash equilibrium and satisfaction level in most experiments, as visualized in Fig. \ref{fig:student_mat}-a, \ref{fig:student_mat}-c. The satisfaction level tends to decrease with the increase in the number of students per group, which is explained with a similar reason as in the real data science dataset. Regarding the fairness w.r.t. protected attribute, the heuristic tends to achieve a higher balance score for the final group assignment in comparison to the knapsack-based method (Fig. \ref{fig:student_mat}-b). When groups' cardinality is low (less than 5), the number of groups generated by the knapsack-based approach is less than the number of groups resulting from the heuristic method (Fig. \ref{fig:student_mat}-d). This phenomenon can be explained by Knapsack's tendency to create groups with a more flexible number of students, which is shown in Fig. \ref{fig:student_mat}-e.
\vspace{-15pt}
\vspace{-17pt}
\subsubsection{UCI student performance - Portuguese}
\label{subsubsec:student_por_result}
\begin{figure*} [!h]
     \centering
     \begin{subfigure}[b]{1\textwidth}
         \centering
         \includegraphics[width=\textwidth]{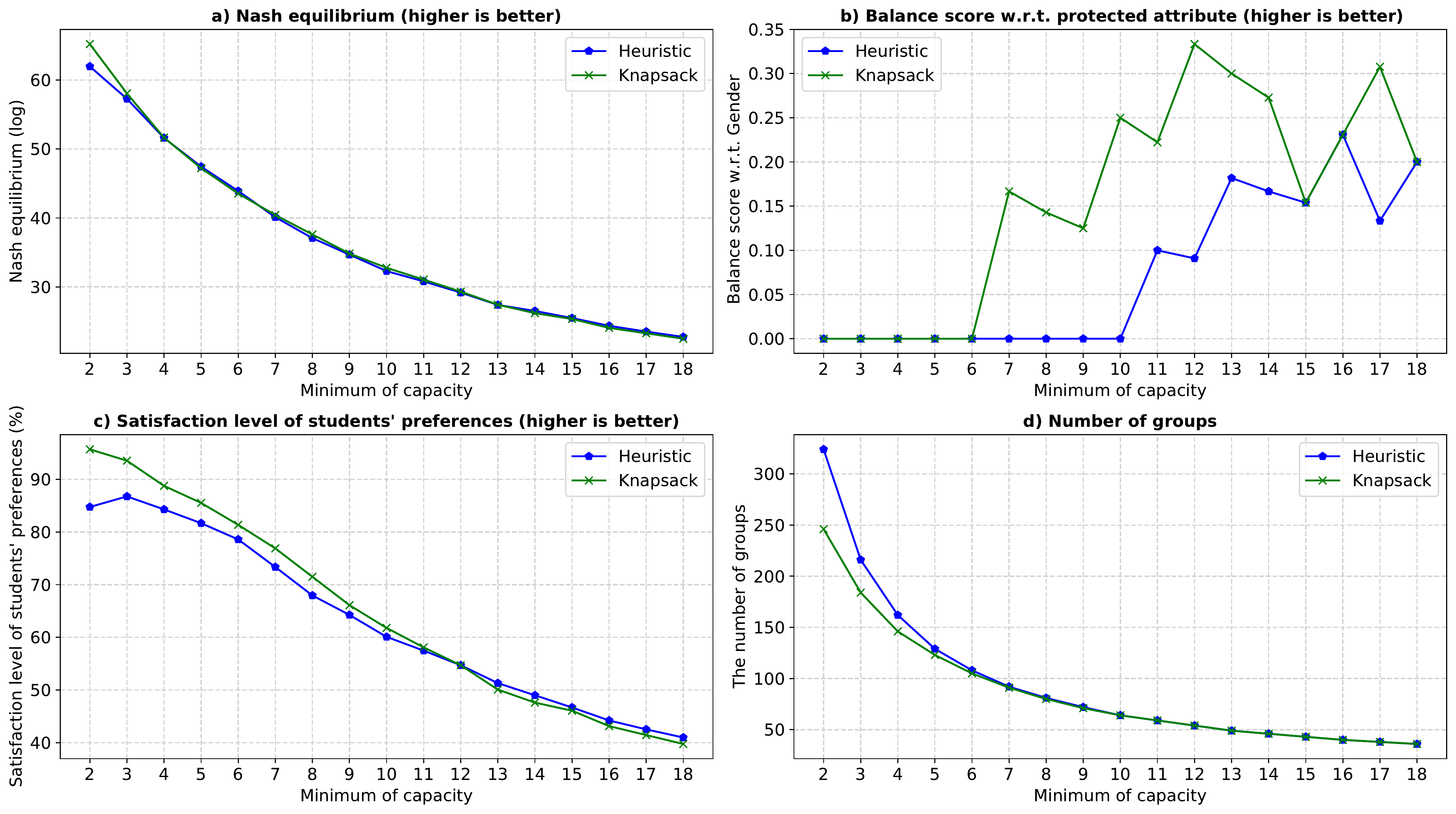}
         \label{fig:plot_student_por}
    \end{subfigure}
     \hfill
     \begin{subfigure}[b]{0.9\textwidth}
         \centering
         \vspace{-13pt}
         \includegraphics[width=\textwidth]{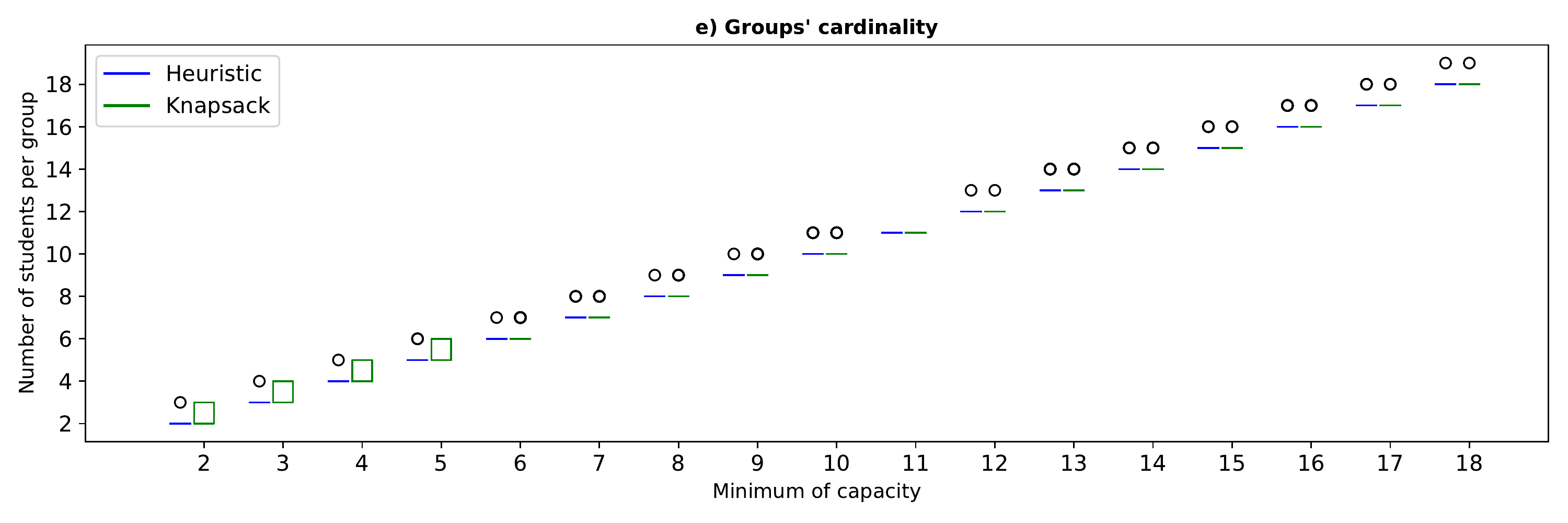}
         \label{fig:boxplot_student_por}
     \end{subfigure}
     \vspace{-17pt}
     \caption{Performance of different methods on the UCI student performance dataset - Portuguese subject}
     \vspace{-10pt}
     \label{fig:student_por}
\end{figure*}

As described in Fig. \ref{fig:student_por}-a and Figure \ref{fig:student_por}-c, the knapsack-based method once again demonstrates the ability to create groups that have higher Nash equilibrium and level of satisfaction w.r.t students' wishes than the heuristic method. In terms of fairness w.r.t gender, a higher balance score is observed in the groups generated by the knapsack-based technique (Fig. \ref{fig:student_por}-b). Regarding the cardinality and the number of groups, similarly to results on the UCI student performance - Mathematics dataset, the knapsack-based approach divides students into more diverse groups in terms of cardinality (Fig. \ref{fig:student_por}-d, e).
\vspace{-15pt}
\vspace{-5pt}

\textbf{Summary of results:} In general, the knapsack-based approach outperforms the heuristic method in terms of the Nash equilibrium, the satisfaction level of students' preferences and fairness w.r.t. gender. However, in some cases, the knapsack-based approach tends to create fewer groups than the heuristic method, i.e., the groups' cardinality is higher, which has both advantages and disadvantages. On the one hand, the larger groups can produce more ideas in brainstorming and discussions \cite{bouchard1970size}. On the other hand, the performance of the group may decline with the increase in the group's size \cite{yetton1983relationships}.
\vspace{-10pt}